\documentclass[conference]{IEEEtran}
\IEEEoverridecommandlockouts
\usepackage{cite}
\usepackage{graphicx}
\usepackage{caption}
\usepackage{hyperref}
\usepackage{booktabs}
\usepackage{url}
\usepackage{amsmath,amssymb,amsfonts}
\usepackage{epsfig}
\usepackage{subfigure}
\usepackage{multirow}
\usepackage{algorithmic}

\usepackage{tabularx}
\usepackage{textcomp}
\usepackage{xcolor}
\usepackage{colortbl}
\usepackage{stfloats}
\def\BibTeX{{\rm B\kern-.05em{\sc i\kern-.025em b}\kern-.08em
    T\kern-.1667em\lower.7ex\hbox{E}\kern-.125emX}}
\begin{document}

\title{From Camera to World: A Plug-and-Play Module for Human Mesh Transformation\\
\thanks{* Corresponding author. This work is supported by the National Natural Science Foundation of China under Grant No. 62072420.}
}

\author{\IEEEauthorblockN{Changhai Ma, Ziyu Wu, Yunkang Zhang, Qijun Ying, Boyan Liu, Xiaohui Cai$^{*}$ }
\IEEEauthorblockA{\textit{ University of Science and Technology of China} \\
\textit{Hefei, China}\\
\{mach1126, wzy1999, ykzhang, yqj, liuboyan\}@mail.ustc.edu.cn, caixiaohui@ustc.edu.cn}
}

\maketitle

\begin{abstract}

Reconstructing accurate 3D human meshes in the world coordinate system from in-the-wild images remains challenging due to the lack of camera rotation information. While existing methods achieve promising results in the camera coordinate system by assuming zero camera rotation, this simplification leads to significant errors when transforming the reconstructed mesh to the world coordinate system. To address this challenge, we propose Mesh-Plug, a plug-and-play module that accurately transforms human meshes from camera coordinates to world coordinates. Our key innovation lies in a human-centered approach that leverages both RGB images and depth maps rendered from the initial mesh to estimate camera rotation parameters, eliminating the dependency on environmental cues. Specifically, we first train a camera rotation prediction module that focuses on the human body's spatial configuration to estimate camera pitch angle. Then, by integrating the predicted camera parameters with the initial mesh, we design a mesh adjustment module that simultaneously refines the root joint orientation and body pose. Extensive experiments demonstrate that our framework outperforms state-of-the-art methods on the benchmark datasets SPEC-SYN and SPEC-MTP.
\end{abstract}


\section{Introduction}
\label{sec:intro}



Reconstructing human meshes from monocular RGB images is an important research task in computer vision. The reconstructed human mesh could empower various human-centered downstream applications like 3D animations, robotics, or AR/VR development. However, inferring 3D information from 2D observations is inherently ill-posed, as multiple 3D configurations can project to the same 2D image. To address this ambiguity, people use large-scale training datasets with 2D joint annotations to train models~\cite{hmr}, integrating human dynamics trees~\cite{tree}, inverse kinematic~\cite{hybrik}, and physical factors~\cite{ipman, physpt} into the human mesh reconstruction task.

\begin{figure}[htbp]
\centerline{\includegraphics[width=1.0\linewidth]{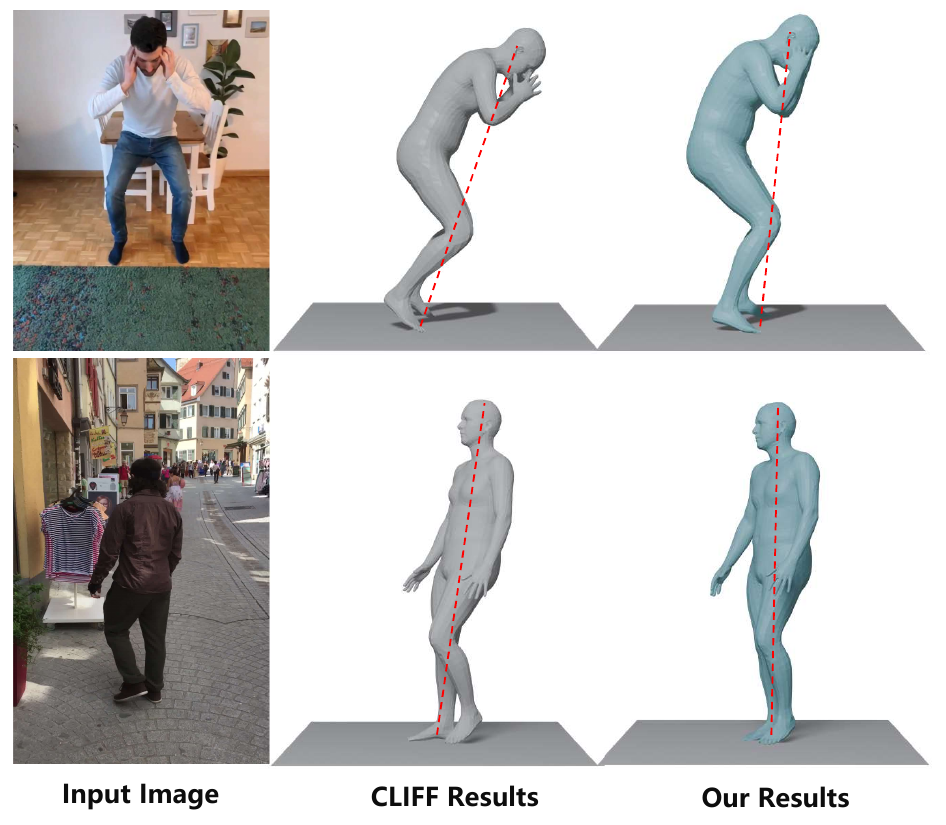}}
\caption{\textbf{Comparison of CLIFF~\cite{cliff} estimation results with our estimation results}. Due to the pitch movement of the camera, the reconstructed results of CLIFF~\cite{cliff} appear tilted. In contrast, our method reconstructs a human mesh that is more consistent with reality.}
\label{fig1}
\end{figure}

Although these works have made significant progress, we observe that most state-of-the-art methods~\cite{hmr,pymaf,cliff,refit}  simplify the camera model. They assumed that the camera rotation was fixed at zero and considered only the camera translation information. This unrealistic assumption leads to satisfactory performance in the camera coordinate system but causes significant errors in the world coordinate system, especially when dealing with angled shots. As shown in Fig~\ref{fig1}, for photos taken from an overhead or low-angle perspective, the reconstructed human mesh tends to exhibit forward or backward leaning. This occurs because assuming zero camera rotation blurs the distinction between body rotation and camera rotation, thereby limiting the model's ability to handle tilted shots effectively.


The main challenge in this task is the inability to obtain accurate camera rotation parameters (pitch, roll and yaw) from monocular images. Although existing works, such as SPEC~\cite{spec}, infer rotation information using pre-trained camera model and apply it for world coordinate pose prediction, this method has its limitations. SPEC~\cite{spec} relies on panoramic image datasets to train its camera rotation estimation model, which performs better in scenes rich in environmental information. However, in human mesh recovery tasks, the human body typically occupies most of the frame, with fewer background environmental details, making SPEC~\cite{spec} less effective for handling such images. Furthermore, estimating camera rotation information from monocular images is inherently difficult, and using imperfect rotation information as input leads to further error accumulation, limiting pose recovery accuracy and weakening the model’s generalization ability.

In order to obtain accurate camera rotation parameters, decouple the body rotation estimation from the camera rotation estimation problem and apply it to reconstruct the human mesh in the world coordinate system, we proposed Mesh-Plug, a \textbf{Plug}-and-Play Module for Human \textbf{Mesh} Transformation, can transform human meshes from the camera coordinate system to the world coordinate system. It consists of two components: the camera rotation estimation module and the mesh adjustment module. Additionally, we introduce a new hybrid loss function to improve performance by differentiating weights of root keypoint and other keypoints.

In the camera rotation estimation module of Mesh-Plug, we focus on a human-centered approach, providing RGB images and depth maps rendered from the 3D human mesh aligned with the camera coordinate system as inputs to the camera rotation estimation module (CamNet). To reduce error accumulation, we only use the camera's pitch angle. This design is based on the human ability to determine whether the viewpoint is from above or below by observing the relative distance between the person and the camera in the image. Therefore, we also leverage the depth map to help the network more accurately estimate the camera's pitch angle. Finally, we combine the initial SMPL~\cite{smpl} parameters, RGB images, and the estimated camera rotation information to obtain the human mesh in the world coordinate system.

In summary, our main contributions are:

\begin{itemize}
  \item We propose Mesh-Plug, a plug-and-play module that effectively transforms human meshes from camera coordinates to world coordinates while adjusting the pose to improve its accuracy in the world coordinate system.
  \item We develop a human-centered approach to estimate camera rotation that leverages depth information and focuses on body configuration, eliminating the dependency on environmental cues.
  \item Experiments on SPEC-SYN~\cite{spec} and SPEC-MTP~\cite{spec} demonstrate that our method significantly improves reconstruction accuracy in the world coordinate system compared to state-of-the-art approaches.
\end{itemize}

\section{Related Work}

\subsection{3D Human Mesh Reconstruction}

Reconstructing a 3D human mesh from a single RGB image is a challenging task. To obtain more accurate and practical human meshes, a parameterized human model, SMPL~\cite{smpl}, has been introduced. This model uses 3D rotations to describe the motion of human joints, which are controlled by predefined linear blend skinning (LBS) weights.

There are two types of mainstream approaches to reconstructing a 3D human mesh from an RGB image: optimization-based methods and regression-based methods. Optimization-based methods~\cite{smplfiy} iteratively fit the body model parameters to 2D evidence through gradient backpropagation. Regression-based methods~\cite{hmr, pymaf, cliff, refit, vibe} directly regress the SMPL parameters or human point clouds through deep neural networks. HMR~\cite{hmr} is the first study to apply regression methods to estimate human model parameters. Since HMR~\cite{hmr}, researchers have extensively explored regression methods, including considering human dynamics trees~\cite{tree}, inverse kinematics~\cite{hybrik}, and physical factors~\cite{ipman, physpt} to improve the accuracy and robustness of the models.

\subsection{Camera Models in Human Mesh Reconstruction Task}

In the 3D human pose recovery task, the application of camera models is crucial for effectively utilizing 2D cues from the image and for visualizing the results. HMR~\cite{hmr} adopts a weak perspective camera model, based on the assumption that the camera's focal length is fixed at 5000 pixels. This assumption is suitable for long-focus images and, due to its simplicity and effectiveness, has been widely adopted in subsequent research~\cite{pymaf,hmr2.0}. To the best of our knowledge, Kissos et al.~\cite{weekly} was the first to improve the weak perspective camera model by not assuming a constant focal length but instead using a constant 55° camera field of view. Later, CLIFF~\cite{cliff}, REFIT~\cite{refit} also adopted this camera model and achieved significant results. In our study, we also use this improved camera model.

Regarding camera rotation, previous works often ignored it, assuming that the camera doesn't rotate. To the best of our knowledge, only SPEC~\cite{spec} and W-HMR~\cite{whmr} focus on the reconstruction of the human mesh in the world coordinate system from single-frame RGB images. SPEC~\cite{spec} is a pioneering work in reconstructing the human mesh in the world coordinate system. It uses CamCalib to predict camera parameters and utilizes these parameters as model inputs to directly reconstruct the human mesh in the world coordinate system. However, SPEC's~\cite{spec} camera parameter prediction relies on environmental information, which does not meet the human-centered task requirements. Additionally, SPEC~\cite{spec} uses excessive and inaccurate camera rotation information as input, leading to error accumulation and reducing reconstruction accuracy. W-HMR~\cite{whmr} continues to use the CamCalib trained by SPEC to reconstruct the human mesh in the world coordinate system.In contrast, our method predicts camera parameters by observing the human body and combines depth maps to assist in estimating more accurate pitch angle information, thus improving the accuracy of the input.

\begin{figure*}[t]
  \centering
   \includegraphics[width=0.95\linewidth]{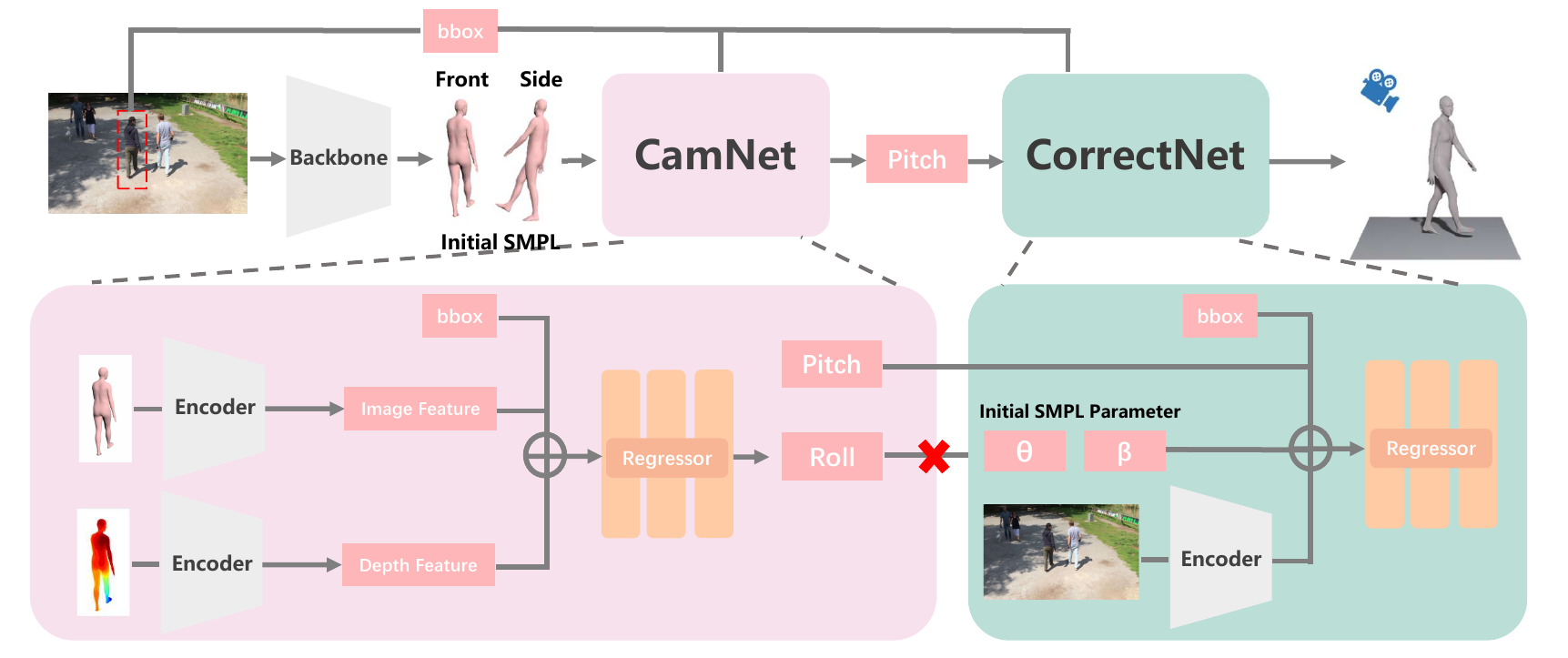}
  \hfill
  \caption{\textbf{Ours pipeline overview}. Given a monocular RGB image and the initial SMPL parameters in the camera coordinate system, we first use the SMPL model to render an RGB image and a depth map from the camera's perspective. These are then input into CamNet to estimate the camera's pitch angle (during training, we estimate the pitch and roll angles, but only the pitch angle is used as input during human mesh reconstruction). Subsequently, the RGB image, initial SMPL parameters, and camera pitch angle are fed into CorrectNet to obtain the human model in the world coordinate system.}
  \label{fig:pipeline}
\end{figure*}

\begin{figure}[htbp]
\centerline{\includegraphics[width=1.0\linewidth]{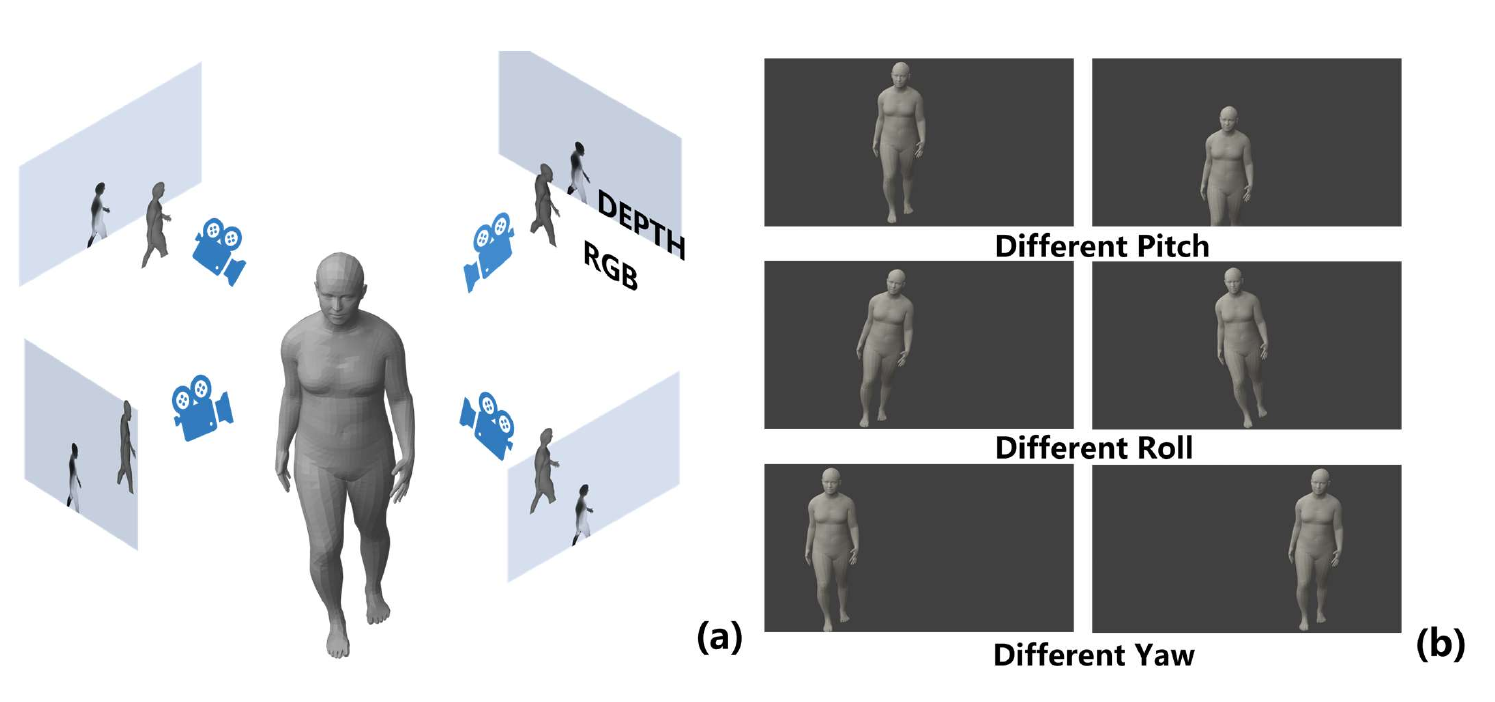}}
\caption{\textbf{(a)} Illustration of the AMASS-Cam data collection process \textbf{(b)} \textbf{From top to bottom}, the effects of different pitch, roll, and yaw angles on the position of the person in the image.}
\label{example}
\end{figure}

\section{Method}

\subsection{SMPL Model}

SMPL~\cite{smpl} is a human parameter model. It provides a formula $M=\mathcal{M}(\theta, \beta)$, where the pose parameters $\theta\in\mathbb{R}^{24\times3}$ and shape parameters $\beta\in\mathbb{R}^{10}$ are input, returning the human mesh model $M\in\mathbb{R}^{N \times 3}$, where the number of vertices $N=6890$. The first three dimensions of the pose parameters $\theta$ represent the root joint orientation, and the remaining 69 dimensions represent the rotation angles of the other joints relative to the root joint. The shape parameters $\beta$ are related to factors such as the person's height and weight.


\subsection{Network Structure}

An overview of the proposed pipeline is shown in Fig~\ref{fig:pipeline}.  The aim of our network is to convert the reconstruction results of the human mesh in the camera coordinate system to the world coordinate system. To accomplish this task, we design the network in two modules: the camera rotation estimation module and the mesh adjustment module. We first pre-train the camera rotation estimation network to predict the pitch angle, which is then used as supplementary information for the mesh adjustment network and in loss computation. This modular design aims to decouple the camera rotation estimation problem from the body orientation problem, enabling each module to focus more effectively on its specific task, thereby improving the accuracy of the reconstruction results. Each module within the network is detailed below.

\subsubsection{CamNet (Camera Rotation Estimation Module)}

CamNet is a human-centered module that can predict the camera's pitch angle without the need for environmental information, as it only requires RGB images and depth maps rendered from the camera's perspective by the SMPL model.

To train CamNet, we create a synthetic dataset, AMASS-Cam, by rendering AMASS~\cite{amass} motion sequences from various camera viewpoints using Blender~\cite{blender}, as illustrated in Fig~\ref{example} (a). The AMASS dataset provides over 40 hours of motion capture data from more than 300 subjects, unified in the SMPL parameter space. We generate over 100,000 training samples by randomly sampling camera positions and rotations, each containing an RGB image, a depth map, and ground truth camera parameters.




CamNet uses ResNet-50~\cite{resnet} , pre-trained on ImageNet~\cite{imgnet} as an encoder to extract features from the RGB image $I$ and the depth map $D$, resulting in $F_i$ and $F_d$, where ${F_i}\in\mathbb{R}^{2048}$ and ${F_d}\in\mathbb{R}^{2048}$. Before being fed into the encoder, these images are cropped and resized to a resolution of $256\times256$ pixels. To enhance the network's ability to perceive the global position of the human subject in the image, we adopt the method from CLIFF~\cite{cliff}, introducing bounding box information $I_{bbox}$. The bounding box information is concatenated with $F_i$ and $F_d$. The combined feature vector is then passed into a regressor to estimate the camera rotation angle.

\begin{equation}
I_{bbox}=\left[\begin{matrix}\dfrac{c_x}{f}&\begin{matrix}\dfrac{c_y}{f}&\dfrac{b}{f}\\\end{matrix}\\\end{matrix}\right],
\end{equation}

where $(c_x,c_y)$ is the position relative to the center of the full image, $b$ is its original size and $f=\sqrt{W^2+H^2}$, $W$ and $H$ represent the width and height of the original image, respectively.


We estimate the pitch and roll angles of the camera simultaneously during the training of CamNet. This is because the camera's rotational information is interdependent; predicting the roll angle can not only help in estimating the pitch angle but also improve the model's understanding of the relative positioning between the camera and the person. We chose not to estimate the yaw angle because its accurate estimation relies on the complete positional information of the person in the image. However, to accommodate images of varying resolutions, we cropped the images before feeding them into the model, which led to the loss of global positional information. Although we implicitly provide the person’s global information through $I_{bbox}$, this is still insufficient to accurately capture the person’s global position, thereby affecting the accuracy of yaw angle estimation.We use $L2$ loss to constrain the training of CamNet.
\begin{equation}
L_{cam}=\lambda_{\alpha}{\Vert \hat\alpha-\alpha \Vert}^2+\lambda_{\gamma}{\Vert \hat\gamma-\gamma \Vert}^2,
\end{equation}

where $\alpha$ and $\gamma$ respectively represent the pitch and roll angles. The hat operator denotes the prediction of that variable. $\lambda's$ are scalar coefficients to balance the loss terms.

\begin{table*}
\centering
\caption{Comparison with state-of-the-art methods on the SPEC-MTP~\cite{spec} and SPEC-SYN~\cite{spec} datasets. 
The best and second-best results are highlighted in bold and underlined, respectively. The first six models in the table disregard camera rotation and output results in camera coordinates. while the last three methods directly provide results in world coordinates.}
\begin{tabularx}{0.9\textwidth}{l|*{3}{>{\centering\arraybackslash}X}|*{3}{>{\centering\arraybackslash}X}} 
\toprule
& \multicolumn{3}{c|}{SPEC-MTP} & \multicolumn{3}{c}{SPEC-SYN} \\ 
\multicolumn{1}{c|}{Models} & WMPJPE$\downarrow$ & PA-MPJPE$\downarrow$ & WPVE$\downarrow$ & WMPJPE$\downarrow$ & PA-MPJPE$\downarrow$ & WPVE$\downarrow$ \\ 
\midrule
GraphCMR~\cite{graphhmr} & 175.1 & 94.3 & 205.5 & 181.7 & 86.6 & 219.8 \\ 
SPIN~\cite{spin} & 143.8 & 79.1 & 165.2 & 165.8 & 79.5 & 194.1 \\ 
PartialHumans~\cite{parti} & 158.9 & 98.7 & 190.1 & 169.3 & 88.2 & 207.6 \\ 
I2L-MeshNet~\cite{i2l} & 167.2 & 99.2 & 199.0 & 169.8 & 82.0 & 203.2 \\ 
HMR~\cite{hmr} & 142.5 & 71.8 & 164.6 & 128.7 & 55.9 & 144.2 \\ 
PyMAF~\cite{pymaf} & 148.8 & 66.7 & 166.7 & 126.8 & \underline{48.7} & 136.7 \\ 
\midrule
SPEC~\cite{spec} & 124.3 & 71.8 & 147.1 & \underline{74.9} & 54.5 & \underline{90.5} \\ 
W-HMR~\cite{whmr} & \underline{118.7} & \underline{66.6} & \underline{133.9} & 82.1 & \textbf{46.1} & 93.3 \\ 
Ours & \textbf{108.1} & \textbf{60.2} & \textbf{127.9} & \textbf{67.5} & \underline{48.7} & 
\textbf{82.0} \\ 
\bottomrule
\end{tabularx}
\label{tab:sota}
\end{table*}

\subsubsection{CorrectNet (Mesh Adjustment Module)}

CorrectNet aims to utilize the camera rotation information obtained from CamNet to reconstruct the human mesh in the world coordinate system.

We choose REFIT~\cite{refit} as the backbone to extract the initial SMPL parameters of the person in the camera coordinate system. First, we use the pre-trained HRNet~\cite{hrnet} to extract image features $F_i$. Then, we concatenate image feature  $F_i$, initial SMPL parameters 
$\theta_c$, $\beta_c$, bounding box information $I_{bbox}$, and pitch angle $P$ (in 6-DoF representation) obtained from CamNet. These combined features are fed into a regressor to predict the SMPL parameters $\theta_w,\beta_w$ and translation $t_b$ in the world coordinate system.
\begin{equation}
\theta_w,\beta_w,t_b=Reg(\oplus(F_i, \theta_c, \beta_c, I_{bbox}, p )).
\end{equation}

We train CorrectNet using the following loss:
\begin{equation}
L_{total}=\lambda_{2D}L_{2D}+\lambda_{3D}L_{3D}+\lambda_{V}L_V+\lambda_{mix}L_{mix}.
\end{equation}

Specifically, we use $L2$ loss to compute the 3D keypoints loss $L_{3D}$ and vertex loss $L_V$
\begin{equation}
L_{3D}={\Vert \hat J_{3D} -J_{3D} \Vert}^2,
\end{equation}
\begin{equation}
L_V={\Vert \hat J_{V·} -J_{V·} \Vert}^2.
\end{equation}

When calculating the 2D ketpoints loss, we introduce the camera rotation to project the predicted 3D keypoints.
\begin{equation}
L_{2D}={\Vert \Pi\hat J_{3D} -J_{2D} \Vert}^2,
\end{equation}

where,
$$\Pi=K\begin{bmatrix} R| -t_b\end{bmatrix},  R=\begin{bmatrix} 1&0&0 \\ 0&cos(P)&-sin(P) \\ 0&sin(P)&cos(P) \end{bmatrix},$$
$$ K=\begin{bmatrix} f&0&W/2 \\ 0&f&H/2 \\ 0&0&1 \end{bmatrix}.$$

Additionally, we introduce a hybrid loss $L_{mix}$ that specifically targets root joint orientation accuracy while maintaining pose coherence:

\begin{equation}
L_{mix}=\lambda_{root}\Vert{\hat\theta_{root}-\theta_{root}}\Vert^2+\Vert\hat\theta-\theta\Vert^2.
\end{equation}

where $\theta_{root}\in\mathbb{R}^{1\times3}$ represents the root joint orientation and $\theta\in\mathbb{R}^{24\times3}$ contains the full pose parameters. 


The reason for this design is that the initial SMPL model obtained from the backbone has two drawbacks: first, the orientation of the root joint exhibits a significant deviation in the world coordinate system; second, due to the similarity of different poses in the projection, there may be ambiguities in the pose prediction of certain body parts.


\begin{figure*}[t]
  \centering
   \includegraphics[width=0.95\linewidth]{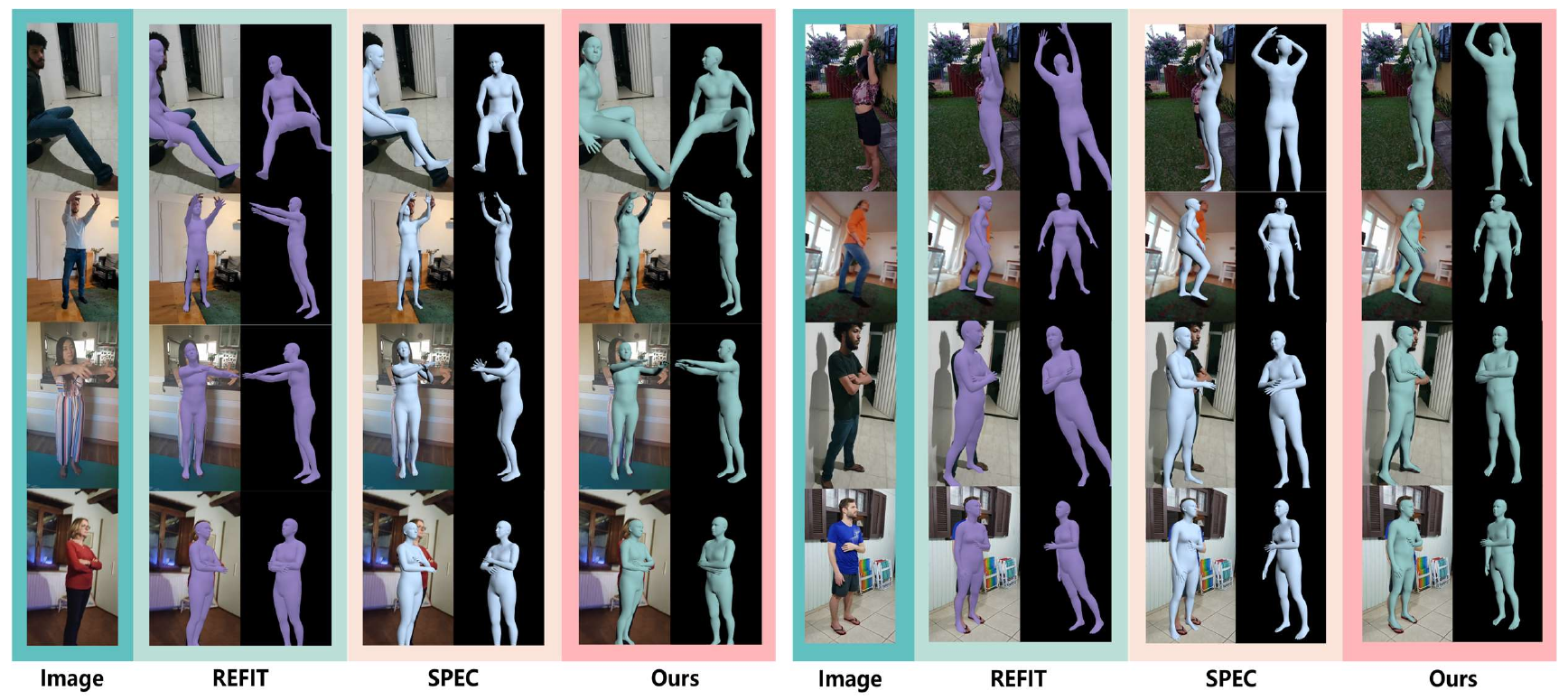}
  \hfill
  \caption{Qualitative comparison between REFIT~\cite{refit}, SPEC~\cite{spec} and Ours on SPEC-MTP~\cite{spec}}
  \label{fig:result}
\end{figure*}

\section{Experiments}

\subsection{Datasets and Evaluation Metrics}\label{AA}
During the training of the CamNet network, we use the AMASS-Cam dataset. For training CorrectNet, we use the SPEC-SYN~\cite{spec} and 3DPW~\cite{3dpw} datasets, both of which provide rich multiview data. In addition, to evaluate the performance of the model, we conduct experiments on the SPEC-SYN~\cite{spec} and SPEC-MTP~\cite{spec} datasets.

MPJPE, PA-MPJPE, and PVE are the most commonly used evaluation metrics in the literature. However, since our work decouples camera rotation from body rotation and directly obtains results in the world coordinate system, we follow the approach of SPEC~\cite{spec} and use variants of MPJPE and PVE, namely W-MPJPE and WPVE, to evaluate the model in the world coordinate system.

\subsection{Implementation Details}

Before feature extraction, the input image is cropped and resized to $256\times256$. During the training of CorrectNet, CamNet is frozen, and no gradient propagation occurs. We use the Adam optimizer with a fixed learning rate of ${5\times10^{-5}}$, and train for 50 epochs on an NVIDIA RTX 4090 GPU.

\begin{table}
\centering
\caption{Effectiveness of Mesh-Plug's Plug-and-play Capability}
\begin{tabular}{lccc}
\toprule
\textbf{Models} & \textbf{WMPJPE$\downarrow$} & \textbf{PA-MPJPE$\downarrow$} & \textbf{WPVE$\downarrow$} \\
\midrule
CLIFF~\cite{cliff} & 137.9 & 64.9 & 157.3 \\
\rowcolor[gray]{0.85} 
CLIFF+Ours & 112.6 & 65.8 & 133.9 \\
REFIT~\cite{refit} & 135.3 & 56.8 & 152.2 \\
\rowcolor[gray]{0.85} 
REFIT+Ours & 108.1 & 60.2 & 127.9 \\
HMR2.0~\cite{hmr2.0} & 146.6 & 70.9 & 155.5 \\
\rowcolor[gray]{0.85} 
HMR2.0+Ours & 110.0 & 62.8 & 128.6 \\
\bottomrule
\end{tabular}
\label{tab:plug}
\end{table}

\subsection{Comparison with State-of-the-art Methods}

We conduct a comprehensive comparative analysis of existing methods on the SPEC-SYN~\cite{spec} and SPEC-MTP~\cite{spec} datasets, evaluating various approaches in both the camera and world coordinate systems. The results are summarized in Table~\ref{tab:sota}. Our proposed method outperforms existing methods across most key metrics in both evaluation datasets. This demonstrates that Mesh-Plug effectively extracts camera rotation information from images and applies it to human pose estimation, significantly improving accuracy in the world coordinate system.

Additionally, we compare our method intuitively with REFIT~\cite{refit} (without considering camera rotation) and SPEC~\cite{spec} (with camera rotation considered), as shown in Fig~\ref{fig:result}. The results highlight the superior performance of our method when dealing with images taken at various angles. Although REFIT~\cite{refit} performs well under the camera coordinate system, it shows significant deviation in predicting the root joint orientation, resulting in noticeable forward leaning of the character. While SPEC~\cite{spec} accurately predicts the root joint orientation in the world coordinate system, it fails to reconstruct the character's movements correctly. In contrast, our method strikes a better balance between accurate motion reconstruction and improves prediction of the root joint orientation.




\begin{table}
\centering
\caption{Effectiveness of Applying Mask Operations to Images}
\begin{tabular}{l|cccc}
\toprule
\multicolumn{1}{c|}{Mask Ratio} & WMPJPE$\downarrow$ & PA-MPJPE$\downarrow$ & WPVE$\downarrow$\\
\midrule
0 & 108.8 & 63.2 & \textbf{127.0} \\
0.2 & \textbf{108.1} & \textbf{60.2} & 127.9 \\
0.5 & 110.6 & 61.3 & 129.2 \\
\bottomrule
\end{tabular}
\label{tab:mask}
\end{table}

\begin{table}
\centering
\caption{Effectiveness of Hybrid Loss Function $L_{mix}$}
\begin{tabular}{l|cccc}
\toprule
\multicolumn{1}{c|}{} & WMPJPE$\downarrow$ & PA-MPJPE$\downarrow$ & WPVE$\downarrow$\\
\midrule
$\lambda_{ori}=0$ & 109.7 & 63.5 & 128.2 \\
$\lambda_{ori}=1$ & 110.2 & 60.6 & 128.9 \\
$\lambda_{ori}=2$ & \textbf{108.1} & \textbf{60.2} & \textbf{127.9} \\
$\lambda_{ori}=3$ & 112.0 & 63.8 & 130.1 \\
\bottomrule
\end{tabular}
\label{tab:ori}
\end{table}


\subsection{Ablation Studies}

In this section, we perform ablation experiments on SPEC-MTP~\cite{spec} to demonstrate the effectiveness of the components.



\subsubsection{Effectiveness of Mesh-Plug's Plug-and-play Capability}

To demonstrate the plug-and-play effectiveness of Mesh-Plug, we selecte three state-of-the-art camera coordinate-based human pose reconstruction methods as backbones to obtain the initial SMPL parameters. Table~\ref{tab:plug} shows that our module significantly improves the performance of human mesh reconstruction methods in the world coordinate system, compared to their performance in the camera coordinate system.

\subsubsection{Effectiveness of Applying Mask Operations to Images}

To enhance the model's generalization ability and its capability to handle occlusions, we apply a mask operation to the image. Before we perform feature extraction on the image,   we divide image into multiple $16\times16$ small squares, and randomly select a certain number of small squares to perform occlusion operations through the mask proportion. To demonstrate the effectiveness of the mask operation, we test different mask ratios, and the best results were obtained with a mask ratio of 0.2. The results are shown in Table~\ref{tab:mask}.

\subsubsection{Effectiveness of Hybrid Loss $L_{mix}$}

We test the impact of different values of $\lambda_{ori}$ in the hybrid loss function, and the results are detailed in Table~\ref{tab:ori}. When $\lambda_{ori}$ was set to 2, we observed the best performance. This confirms the effectiveness of the hybrid loss function we designed, as it allows the network to allocate different attention weights to the root joint orientation and the overall joint parameters, thereby optimizing the model's performance.

\section*{Conclusion}


In this paper, we introduce Mesh-Plug, a plug-and-play module designed to transform human mesh reconstruction results from the camera coordinate system to the world coordinate system. Through a modular design, Mesh-Plug decouples camera rotation estimation from the problem of mesh reconstruction in the world coordinate system. Additionally, we propose a human-centered camera rotation prediction method that does not rely on environmental cues. Extensive experiments demonstrate that our approach advances the state-of-the-art performance on benchmark datasets.



\bibliographystyle{IEEEbib}
\bibliography{icme2025references}

@article{hmr,
  title={End-to-End Recovery of Human Shape and Pose},
  author={Angjoo Kanazawa and Michael J. Black and David W. Jacobs and Jitendra Malik},
  journal={2018 IEEE/CVF Conference on Computer Vision and Pattern Recognition (CVPR)},
  year={2017},
  pages={7122-7131},
  url={https://api.semanticscholar.org/CorpusID:28772744}
}

@article{vibe,
  title={VIBE: Video Inference for Human Body Pose and Shape Estimation},
  author={Muhammed Kocabas and Nikos Athanasiou and Michael J. Black},
  journal={2020 IEEE/CVF Conference on Computer Vision and Pattern Recognition (CVPR)},
  year={2019},
  pages={5252-5262},
  url={https://api.semanticscholar.org/CorpusID:209323955}
}

@article{pymaf,
  title={PyMAF: 3D Human Pose and Shape Regression with Pyramidal Mesh Alignment Feedback Loop},
  author={Hongwen Zhang and Yating Tian and Xinchi Zhou and Wanli Ouyang and Yebin Liu and Limin Wang and Zhenan Sun},
  journal={2021 IEEE/CVF International Conference on Computer Vision (ICCV)},
  year={2021},
  pages={11426-11436},
  url={https://api.semanticscholar.org/CorpusID:232417104}
}

@inproceedings{cliff,
  title={CLIFF: Carrying Location Information in Full Frames into Human Pose and Shape Estimation},
  author={Zhihao Li and Jianzhuang Liu and Zhensong Zhang and Songcen Xu and Youliang Yan},
  booktitle={European Conference on Computer Vision (ECCV)},
  year={2022},
  url={https://api.semanticscholar.org/CorpusID:251224390}
}

@article{refit,
  title={ReFit: Recurrent Fitting Network for 3D Human Recovery},
  author={Yufu Wang and Kostas Daniilidis},
  journal={2023 IEEE/CVF International Conference on Computer Vision (ICCV)},
  year={2023},
  pages={14598-14608},
  url={https://api.semanticscholar.org/CorpusID:261065129}
}

@article{spec,
  title={SPEC: Seeing People in the Wild with an Estimated Camera},
  author={Muhammed Kocabas and Chun-Hao Paul Huang and J. Tesch and Lea Muller and Otmar Hilliges and Michael J. Black},
  journal={2021 IEEE/CVF International Conference on Computer Vision (ICCV)},
  year={2021},
  pages={11015-11025},
  url={https://api.semanticscholar.org/CorpusID:238260044}
}

@article{smpl,
  title={SMPL: A Skinned Multi-Person Linear Model},
  author={Matthew Loper and Naureen Mahmood and Javier Romero and Gerard Pons-Moll and Michael J. Black},
  journal={Seminal Graphics Papers: Pushing the Boundaries, Volume 2},
  year={2023},
  url={https://api.semanticscholar.org/CorpusID:5328073}
}

@article{smplfiy,
  title={Keep It SMPL: Automatic Estimation of 3D Human Pose and Shape from a Single Image},
  author={Federica Bogo and Angjoo Kanazawa and Christoph Lassner and Peter Gehler and Javier Romero and Michael J. Black},
  journal={ArXiv},
  year={2016},
  volume={abs/1607.08128},
  url={https://api.semanticscholar.org/CorpusID:13438951}
}

@article{tree,
  title={Encoder-decoder with Multi-level Attention for 3D Human Shape and Pose Estimation},
  author={Ziniu Wan and Zhengjia Li and Maoqing Tian and Jianbo Liu and Shuai Yi and Hongsheng Li},
  journal={2021 IEEE/CVF International Conference on Computer Vision (ICCV)},
  year={2021},
  pages={13013-13022},
  url={https://api.semanticscholar.org/CorpusID:237421205}
}

@article{HybrIK,
  title={HybrIK: A Hybrid Analytical-Neural Inverse Kinematics Solution for 3D Human Pose and Shape Estimation},
  author={Jiefeng Li and Chao Xu and Zhicun Chen and Siyuan Bian and Lixin Yang and Cewu Lu},
  journal={2021 IEEE/CVF Conference on Computer Vision and Pattern Recognition (CVPR)},
  year={2020},
  pages={3382-3392},
  url={https://api.semanticscholar.org/CorpusID:227228054}
}

@article{ipman,
  title={3D Human Pose Estimation via Intuitive Physics},
  author={Shashank Tripathi and Lea Muller and Chun-Hao Paul Huang and Omid Taheri and Michael J. Black and Dimitrios Tzionas},
  journal={2023 IEEE/CVF Conference on Computer Vision and Pattern Recognition (CVPR)},
  year={2023},
  pages={4713-4725},
  url={https://api.semanticscholar.org/CorpusID:257900989}
}

@article{physpt,
  title={PhysPT: Physics-aware Pretrained Transformer for Estimating Human Dynamics from Monocular Videos},
  author={Yufei Zhang and Jeffrey O. Kephart and Zijun Cui and Qiang Ji},
  journal={2024 IEEE/CVF Conference on Computer Vision and Pattern Recognition (CVPR)},
  year={2024},
  pages={2305-2317},
  url={https://api.semanticscholar.org/CorpusID:269004769}
}

@article{hmr2.0,
  title={Humans in 4D: Reconstructing and Tracking Humans with Transformers},
  author={Shubham Goel and Georgios Pavlakos and Jathushan Rajasegaran and Angjoo Kanazawa and Jitendra Malik},
  journal={2023 IEEE/CVF International Conference on Computer Vision (ICCV)},
  year={2023},
  pages={14737-14748},
  url={https://api.semanticscholar.org/CorpusID:258987755}
}

@inproceedings{weekly,
  title={Beyond Weak Perspective for Monocular 3D Human Pose Estimation},
  author={Imry Kissos and Lior Fritz and Matan Goldman and Omer Meir and Eduard Oks and Mark Kliger},
  booktitle={ECCV Workshops},
  year={2020},
  url={https://api.semanticscholar.org/CorpusID:221655480}
}

@manual{blender,
    title={Blender},
    note="\url{http://www.blender.org}",
    year="2023"
}

@article{amass,
  title={AMASS: Archive of Motion Capture As Surface Shapes},
  author={Naureen Mahmood and Nima Ghorbani and Nikolaus F. Troje and Gerard Pons-Moll and Michael J. Black},
  journal={2019 IEEE/CVF International Conference on Computer Vision (ICCV)},
  year={2019},
  pages={5441-5450},
  url={https://api.semanticscholar.org/CorpusID:102351100}
}

@article{imgnet,
  title={ImageNet: A large-scale hierarchical image database},
  author={Jia Deng and Wei Dong and Richard Socher and Li-Jia Li and K. Li and Li Fei-Fei},
  journal={2009 IEEE Conference on Computer Vision and Pattern Recognition},
  year={2009},
  pages={248-255},
  url={https://api.semanticscholar.org/CorpusID:57246310}
}

@article{resnet,
  title={Deep Residual Learning for Image Recognition},
  author={Kaiming He and X. Zhang and Shaoqing Ren and Jian Sun},
  journal={2016 IEEE Conference on Computer Vision and Pattern Recognition (CVPR)},
  year={2015},
  pages={770-778},
  url={https://api.semanticscholar.org/CorpusID:206594692}
}

@article{graphhmr,
  title={Convolutional Mesh Regression for Single-Image Human Shape Reconstruction},
  author={Nikos Kolotouros and Georgios Pavlakos and Kostas Daniilidis},
  journal={2019 IEEE/CVF Conference on Computer Vision and Pattern Recognition (CVPR)},
  year={2019},
  pages={4496-4505},
  url={https://api.semanticscholar.org/CorpusID:119064973}
}

@article{spin,
  title={Learning to Reconstruct 3D Human Pose and Shape via Model-Fitting in the Loop},
  author={Nikos Kolotouros and Georgios Pavlakos and Michael J. Black and Kostas Daniilidis},
  journal={2019 IEEE/CVF International Conference on Computer Vision (ICCV)},
  year={2019},
  pages={2252-2261},
  url={https://api.semanticscholar.org/CorpusID:203591723}
}

@article{parti,
  title={Full-Body Awareness from Partial Observations},
  author={C. Rockwell and David F. Fouhey},
  journal={ArXiv},
  year={2020},
  volume={abs/2008.06046},
  url={https://api.semanticscholar.org/CorpusID:221112129}
}

@article{i2l,
  title={I2L-MeshNet: Image-to-Lixel Prediction Network for Accurate 3D Human Pose and Mesh Estimation from a Single RGB Image},
  author={Gyeongsik Moon and Kyoung Mu Lee},
  journal={ArXiv},
  year={2020},
  volume={abs/2008.03713},
  url={https://api.semanticscholar.org/CorpusID:221090782}
}

@inproceedings{whmr,
  title={W-HMR: Monocular Human Mesh Recovery in World Space with Weak-Supervised Calibration},
  author={Wei Yao and Hongwen Zhang and Yunlian Sun and Jinhui Tang},
  year={2023},
  url={https://api.semanticscholar.org/CorpusID:265498801}
}

@article{hrnet,
  title={Deep High-Resolution Representation Learning for Visual Recognition},
  author={Jingdong Wang and Ke Sun and Tianheng Cheng and Borui Jiang and Chaorui Deng and Yang Zhao and Dong Liu and Yadong Mu and Mingkui Tan and Xinggang Wang and Wenyu Liu and Bin Xiao},
  journal={IEEE Transactions on Pattern Analysis and Machine Intelligence},
  year={2019},
  volume={43},
  pages={3349-3364},
  url={https://api.semanticscholar.org/CorpusID:201124533}
}

@inproceedings{3dpw,
  title={Recovering Accurate 3D Human Pose in the Wild Using IMUs and a Moving Camera},
  author={Timo von Marcard and Roberto Henschel and Michael J. Black and Bodo Rosenhahn and Gerard Pons-Moll},
  booktitle={European Conference on Computer Vision},
  year={2018},
  url={https://api.semanticscholar.org/CorpusID:273160019}
}

\end{document}